\documentclass[review]{elsarticle}
\usepackage{geometry}
\usepackage{lineno,hyperref}
\modulolinenumbers[5]
\usepackage{algorithm}
\usepackage{algorithmic}
\usepackage{amssymb}
\usepackage{amsmath}
\usepackage{mathrsfs}
\usepackage{graphicx,graphics,subfigure}
\usepackage{amsfonts}
\usepackage{amsmath}
\usepackage{dcolumn}
\usepackage{bigstrut}
\usepackage{pdfpages}
\usepackage{epstopdf}
\usepackage{color}
\usepackage{booktabs}
\usepackage{url}
\usepackage{multirow}
\usepackage{multicol}
\usepackage{algorithm}
\usepackage{algorithmic}
\journal{Signal Processing}

\begin{document}

\begin{frontmatter}

\title{Bayesian Fusion for Infrared and Visible Images}

%
%
%
\author{Zixiang Zhao, Shuang Xu, Chunxia Zhang, Junmin Liu, and Jiangshe Zhang\corref{mycorrespondingauthor}}
\address{School of Mathematics and Statistics, Xi'an Jiaotong University, China}
\fntext[myfootnote]{E-mail: zixiangzhao@stu.xjtu.edu.cn, shuangxu@stu.xjtu.edu.cn,  cxzhang@mail.xjtu.edu.cn, junminliu@mail.xjtu.edu.cn, jszhang@mail.xjtu.edu.cn.}

\cortext[mycorrespondingauthor]{Corresponding author}
\begin{abstract}
Infrared and visible image fusion has been a hot issue in image fusion. In this task, a fused image containing both the gradient and detailed texture information of visible images as well as the thermal radiation and highlighting targets of infrared images is expected to be obtained. In this paper, a novel Bayesian fusion model is established for infrared and visible images. In our model, the image fusion task is cast into a regression problem. To measure the variable uncertainty, we formulate the model in a hierarchical Bayesian manner. Aiming at making the fused image satisfy human visual system, the model incorporates the total-variation(TV) penalty.
Subsequently, the model is efficiently inferred by the expectation-maximization(EM) algorithm. We test our algorithm on TNO and NIR image fusion datasets with several state-of-the-art approaches.
Compared with the previous methods, the novel model can generate better fused images with high-light targets and rich texture details, which can improve the reliability of the target automatic detection and recognition system.
\end{abstract}

\begin{keyword}
image fusion\sep hierarchical Bayesian model\sep total-variation penalty\sep EM algorithm
\end{keyword}

\end{frontmatter}
\section{Introduction}
Image fusion, as an information-enhanced image processing, is a hot issue in computer vision today.
Image fusion is an enhancement image processing technique to produce a robust or informative image\cite{ma2019infrared}. Image fusion has a wide range of applications in pattern recognition \cite{singh2008integrated}, medical imaging\cite{zong2017medical}, remote sensing\cite{simone2002image}, and modern military\cite{chen2014image} as they require to fuse two or more images in  same scenes\cite{li2017pixel}.

The fusion of visible and infrared images improves the perception ability of human visual system in target detection and recognition\cite{kong2007multiscale}.
As we know, a visible image has rich appearance information, and the features such as texture and detail information are often not obvious in the corresponding infrared image. In contrast, an infrared image mainly reflects the heat radiation emitted by objects, which is less affected by illumination changes or artifacts and overcome the obstacles to target detection at night. However, the spatial resolution of infrared images is typically lower than that of visible images. Consequently, fusing thermal radiation and texture detail information into an image facilitates automatic detection and accurate positioning of targets\cite{ma2016infrared}.

Broadly speaking, the current algorithms for fusing visible and infrared images can be divided into four categories: multi-scale transformation, sparse representation, subspace and saliency methods\cite{ma2019infrared}.
The multi-scale transformation based methods\cite{liu2018deep,li2011performance,pajares2004wavelet, zhang1999categorization}, in general, decompose source images into multiple levels and then fuse images from the same level of the decomposed layers in specific fusion strategies. Finally, the fused image is recovered by incorporating the fused layers.
The second category is sparse representation-based methods \cite{yang2014visual,wang2014fusion, li2012group}, which assume that the natural image is a sparse linear combination of itself, and fused images can be recovered by merging the coefficients.
The third category is the subspace learning-based methods\cite{bavirisetti2017multi,kong2014adaptive,patil2011image}, which aims to project high-dimensional input images into low-dimensional subspaces to capture the intrinsic feature of the original image.
The fourth category is saliency-based methods\cite{bavirisetti2016two, zhang2017infrared, zhao2014infrared}. Based on the prior knowledge that humans usually pay more attention to the saliency objects rather than surrounding areas, they fuse images by maintaining the integrity of the salient target areas.

To the best of our knowledge, no Bayesian model has been applied to the image fusion problem.
Therefore, we present in this paper a novel Bayesian fusion model for infrared and visible images. In our model, the image fusion task is cast into a regression problem. To measure the variable uncertainty, we formulate the model in the hierarchical Bayesian manner. Besides, to make the fused image satisfy human visual system, the model incorporates the TV penalty. 
Then, this model is efficiently inferred by EM algorithm. We test our algorithm on TNO and NIR image fusion datasets with several state-of-the-art approaches.
Compared with the previous methods, this method can generate fused image results with high-light targets and rich texture details, which can improve the reliability of the target automatic detection and recognition system.

The rest paper is organized as follows.
In section \ref{sec:1}, we introduced the Bayesian fusion method. In section \ref{sec:2}, some experiments are conducted to investigate and compare the proposed method with some state-of-the-art techniques. Finally, some conclusions are drawn in section \ref{sec:3}.

\section{Bayesian fusion model}\label{sec:1}
In this section, we present a novel Bayesian fusion model for infrared and visible images. Then, this model is efficiently inferred by the EM algorithm\cite{dempster1977maximum}.
\subsection{Model formulation}
Given a pair of pre-registered infrared and visible images, $U,V\in \mathbb{R}^{h\times w}$, image fusion technique aims at obtaining an informative image $I$ from $U$ and $V$.

It is well-known that visible images satisfy human visual perception, while they are significantly sensitive to disturbances, such as poor illumination, fog and so on. In contrast, infrared images are robust to these disturbances but may lose part of informative textures. In order to preserve the general profile of two images, we minimize the difference between fused and source images, that is
$$
	\min_{I} f(U,I)+g(V,I),
$$
where $f,g$ are loss functions. Typically, we assume the difference is measured by $L_1$ norm. Thus, the problem can be rewritten as
$$\min_{I} ||I-U||_1+||I-V||_1.$$  Let $X=I-V$ and $Y=U-V$, then we have
\begin{equation}\label{eq:reg}
	\min_{X} ||X-Y||_1+||X||_1.
\end{equation}
Essentially, equation (\ref{eq:reg}) corresponds to a linear regression model
\begin{equation*}
	Y=X+E,
\end{equation*}
where $E$ denotes a Laplacian noise and $X$ is governed by Laplacian distribution. By reformulating this problem in the Bayesian fashion, the conditional distribution of $Y$ given $X$ is
\begin{equation*}\label{eq:prior_y}
\begin{aligned}
	p(Y|X) &= {\rm Laplace} (Y|X,\lambda_y) \\
	&=\prod_{i,j} \frac{1}{2\lambda_y} \exp\left(-\frac{|y_{ij}-x_{ij}|}{\lambda_y}\right),
\end{aligned}
\end{equation*}
and the prior distribution of $X$ is
\begin{equation*}\label{eq:prior_x}
\begin{aligned}
p(X) &= {\rm Laplace}(X|0,\tau_x) \\
&= \prod_{i,j} \frac{1}{2\tau_x} \exp\left(-\frac{|x_{ij}|}{\tau_x}\right).
\end{aligned}
\end{equation*}
To avert from $L_1$ norm optimization, we reformulate Laplacian distribution as Gaussian scale mixtures with exponential distributed prior to the variance, that is,
\begin{equation}\label{eq:lap2normal}
\begin{aligned}
&{\rm Laplace}(\xi|\mu,\sqrt{\lambda/2}) \\
=& \frac{1}{2}\sqrt{\frac{2}{\lambda}} \exp\left(-\sqrt{\frac{2}{\lambda}} |\xi-\mu|\right) \\
=& \int_{0}^{\infty}\frac{1}{\sqrt{2\pi a}} \exp\left(-\frac{(\xi-\mu)^2}{2a}\right)\frac{1}{\lambda}\exp\left(-\frac{a}{\lambda}\right) da\\
=& \int_{0}^{\infty} \mathcal{N}(\xi|\mu,a) {\rm Expoential}(a|\lambda) da
\end{aligned}
\end{equation}
where $\mathcal{N}(\xi|\mu,a)$ denotes Gaussian distribution with mean $\mu$ and variance $a$, and ${\rm Exp}(a|\lambda)$ denotes exponential distribution with scale parameter $\lambda$. According to equation (\ref{eq:lap2normal}), the original model of $p(Y|X)$ and $p(X)$ can be rewritten in the hierarchical Bayesian manner, that is,
\begin{equation*}
\begin{cases}
y_{ij}|x_{ij},a_{ij} \sim \mathcal{N}(y_{ij}|x_{ij},a_{ij}) \\
a_{ij} \sim {\rm Exp}(a_{ij}|\lambda)\\
x_{ij}|b_{ij} \sim \mathcal{N}(x_{ij}|0,b_{ij}) \\
b_{ij} \sim {\rm Exp}(b_{ij}|\tau),
\end{cases}
\end{equation*}
for all $i=1,\cdots,h$ and $j=1,\cdots,w$, where $h$ and $w$ mean the height and the width of the input image. In what follows, we use matrices $A$ and $B$ to denote the collection of all latent variables $a_{ij}$ and $b_{ij}$, respectively.

Besides modeling the general profiles, the image textures should be taken into consideration so as to make fused image satisfy the human visual perception. As discussed above, there is plenty of high-frequency information in visible images, but the corresponding areas often cannot be observed in infrared images. In order to preserve the edge information of visible images, we regularize the fused image in gradient domain with a gradient sparsity regularizer expressed as
\begin{equation*}
	h(X) = \frac{1}{2}\lambda_g ||\nabla I-\nabla V||_1=\frac{1}{2}\lambda_g ||\nabla X||_1,
\end{equation*}
where $\lambda_g$ is a hyper-parameter controlling the strength of regularization, $\nabla$ denotes the gradient operator. This regularizer makes the fused image have similar textures to the visible image.

\begin{figure}
	\centering
	\includegraphics[width=0.7\linewidth]{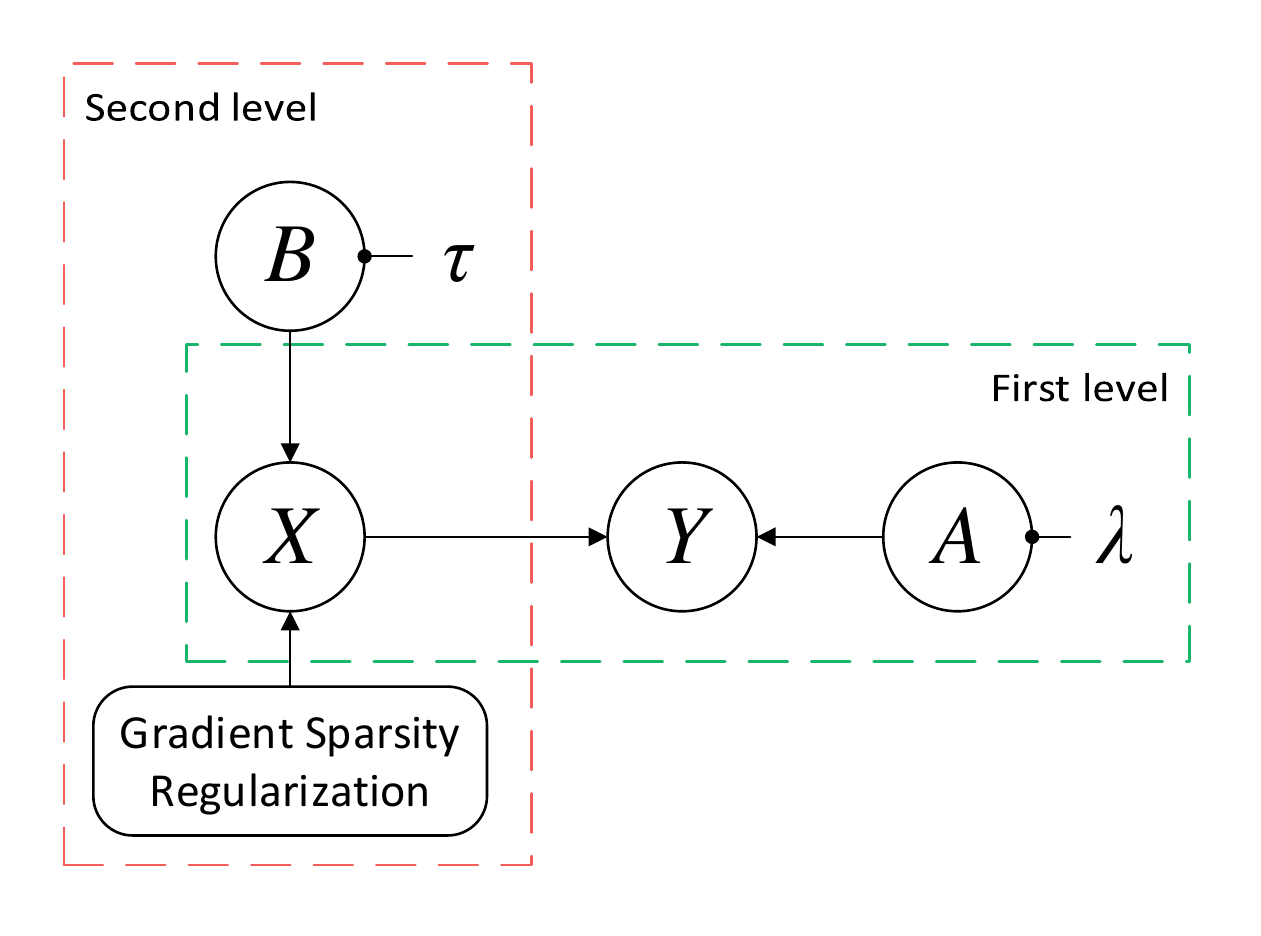}
\vspace{-1em}
	\caption{Illustration of our Bayesian graph model.}
\vspace{-1em}
	\label{fig:bayesmodel}
\end{figure}

By combining general profiles and gradients modeling, Fig. \ref{fig:bayesmodel} displays the graphical expression of our hierarchical Bayesian model. Specifically, in the first level, $A$ is a latent variable, while $Y$ and $X$ are observed and unknown variables, respectively. In the second level, $X$ and $B$ are latent and observed variables, respectively. And $\lambda$ and $\tau$ are hyper-parameters. By ignoring the constant not depending on $X$, the log-likelihood of the model can be expressed as
\begin{equation*}
\begin{aligned}
	\ell(X)& = \log p(Y,X)-h(X)\\
	& = -\sum_{i,j}\left[\frac{(x_{ij}-y_{ij})^2}{2a_{ij}} + \frac{x_{ij}^2}{2b_{ij}}\right] - \frac{\lambda_g}{2}||\nabla X||_1.
\end{aligned}
\end{equation*}
In next subsection, we will discuss how to infer this model.

\subsection{Model inference}
As is well-known, the EM algorithm is an effective tool to maximize the log-likelihood function of a problem which involves some latent variables. In detail, we firstly initialize unknown variable $X$. Then, in E-step, it calculates the expectation of log-likelihood function with respect to $p(A,B|X^{(t)},Y)$, which is often referred to as the so-called $Q$-function,
$$
Q(X|X^{(t)}) = \mathbb{E}_{A,B|X^{(t)},Y}[\ell(X)].
$$
In M-step, we find $X$ to maximize the $Q$-function, i.e.,
$$
X^{(t+1)} = \arg\max_{X} Q(X|X^{(t)}).
$$

\noindent\textbf{E-step}: In order to obtain the $Q$-function in our model, $\mathbb{E}_{a_{ij}|x_{ij}^{(t)},y_{ij}}[1/a_{ij}]$ and $\mathbb{E}_{b_{ij}|x_{ij}^{(t)}}[1/b_{ij}]$ should be computed. For convenience, we had better compute the posterior distribution for $\tilde{a}_{ij}\equiv1/a_{ij}$ and $\tilde{b}_{ij}\equiv 1/b_{ij}$. It has been assumed that the prior distribution of $a_{ij}$ is ${\rm Exp}(a_{ij}|\lambda)$, so $\tilde{a}_{ij}$ is governed by \textit{inverse gamma distribution} with shape parameter of 1 and scale parameter of $1/\lambda$. And the probability density function of $\tilde{a}$ is given by
\begin{equation*}
	p(\tilde{a}) = \frac{1}{\lambda} \tilde{a}^{-2}\exp\left(-\frac{1}{\lambda\tilde{a}}\right)
\end{equation*}
According to the Bayesian formula, the posterior of $\tilde{a}_{ij}$ is the \textit{inverse Gaussian distribution}, that is,
\begin{equation*}
\begin{aligned}
&	\log p(\tilde{a}_{ij}|y_{ij},x_{ij}) \\
=& \log p(y_{ij}|x_{ij},a_{ij})+\log p(\tilde{a}_{ij})\\
=& -\frac{\log a_{ij}}{2} - \frac{(y_{ij}-x_{ij})^2}{2a_{ij}} -2\log\tilde{a}_{ij} -\frac{1}{\lambda\tilde{a}_{ij}}+{\rm constant}\\
=& \frac{\log \tilde{a}_{ij}}{2} - \frac{\tilde{a}_{ij}(y_{ij}-x_{ij})^2}{2} -2\log\tilde{a}_{ij} -\frac{1}{\lambda\tilde{a}_{ij}}+{\rm constant}\\
=&-\frac{3}{2}\log \tilde{a}_{ij}- \frac{\tilde{a}_{ij}(y_{ij}-x_{ij})^2}{2}-\frac{1}{\lambda\tilde{a}_{ij}}+{\rm constant}.
\end{aligned}
\end{equation*}
where $\alpha_{ij}=\sqrt{2(y_{ij}-x_{ij})^2/\lambda}$ and $\tilde{\lambda}=2/\lambda$. As for $\tilde{b}_{ij}$, we can compute its posterior in the same way,
\begin{equation*}
\begin{aligned}
&	\log p(\tilde{b}_{ij}|x_{ij}) \\
=& \log p(x_{ij}|b_{ij})+\log p(\tilde{b}_{ij})\\
=& -\frac{\log b_{ij}}{2} - \frac{x_{ij}^2}{2b_{ij}} -2\log\tilde{b}_{ij} -\frac{1}{\tau\tilde{b}_{ij}}+{\rm constant}\\
=&-\frac{3}{2}\log \tilde{b}_{ij}- \frac{\tilde{b}_{ij}x_{ij}^2}{2}-\frac{1}{\tau\tilde{b}_{ij}}+{\rm constant}.
\end{aligned}
\end{equation*}
Similarly, the posterior of $\tilde{b}_{ij}$ is
\begin{equation*}
p(\tilde{b}_{ij}|x_{ij}) = \mathcal{IN}(\tilde{b}_{ij}|\beta_{ij},\tilde{\tau}),
\end{equation*}
where $\beta_{ij}=\sqrt{2x_{ij}^2/\tau}$ and $\tilde{\tau}=2/\tau$. Note that the expectation of inverse Gaussian distribution is its location parameter. Thus, we have
\begin{equation}\label{eq:E-step1}
\mathbb{E}_{a_{ij}|x_{ij}^{(t)},y_{ij}}\left[\frac{1}{a_{ij}}\right] = \alpha_{ij}=\sqrt{\frac{2(y_{ij}-x_{ij}^{(t)})^2}{\lambda}} ,
\end{equation}
\begin{equation}\label{eq:E-step2}
\mathbb{E}_{b_{ij}|x_{ij}^{(t)}}\left[\frac{1}{b_{ij}}\right] = \sqrt{\frac{2[x_{ij}^{(t)}]^2}{\tau}} .
\end{equation}
Thereafter, in E-step, the $Q$-function is given by
\begin{equation*}
\begin{aligned}
Q &=-\sum_{i,j}\left[\frac{1}{2}\alpha_{ij}(x_{ij}-y_{ij})^2 +\frac{1}{2}\beta_{ij} x_{ij}^2\right] - \frac{1}{2}\lambda_g||\nabla X||_1 \\
&=-||W_1\odot(X-Y)||_2^2-||W_2\odot X||_2^2- \lambda_g||\nabla X||_1.
\end{aligned}
\end{equation*}
where the symbol $\odot$ means element-wise multiplication, and the $(i,j)$th entries of $W_1$ and $W_2$ are $\sqrt{\alpha_{ij}}$ and $\sqrt{\beta_{ij}}$, respectively.

\noindent\textbf{M-step}: Here, we need to minimize the negative $Q$-function with respect to $X$. The half-quadratic splitting algorithm is employed to deal with this problem, i.e.,
\begin{equation*}
\begin{aligned}
	&\min_{X,F,H} ||W_1\odot(X-Y)||_2^2+||W_2\odot X||_2^2+ \lambda_g||F||_1, \\
	&{\rm s.t.\quad} F=\nabla H, H=X.
\end{aligned}
\end{equation*}
It can be further cast into the following unconstraint optimization problem,
\begin{equation*}
\begin{aligned}
\min_{X,F,H} & ||W_1\odot(X-Y)||_2^2+||W_2\odot X||_2^2+ \lambda_g||F||_1  \\
 &+ \frac{\rho}{2}\left( ||F-\nabla H||_2^2+||H-X||_2^2 \right) .
\end{aligned}
\end{equation*}
The unknown variables $X,F,H$ can be solved iteratively in the coordinate descent fashion.

\noindent\textbf{Update $X$}: It is a least squares issue,
\begin{equation*}
\min_{X}	||W_1\odot(X-Y)||_2^2+||W_2\odot X||_2^2+\frac{\rho}{2}||H-X||_2^2.
\end{equation*}
The solution of $X$ is
\begin{equation} \label{eq:X}
	X = (2W_1^2\odot Y+\rho H) \oslash (2W_1^2+2W_2^2+\rho),
\end{equation}
where the symbol $\oslash$ means the element-wise division.

\noindent\textbf{Update $F$}: It is an $L_1$ norm penalized regression issue,
\begin{equation*}
	\min_{F} \lambda||F||_1+\frac{\rho}{2} ||F-\nabla H||_2^2.
\end{equation*}
The solution is
\begin{equation}\label{eq:F}
F = S(\nabla H,\lambda/\rho),
\end{equation}
where $S(x,\gamma) = {\rm sign}(x)\max(|x|-\gamma,0)$.

\noindent\textbf{Update $H$}: It is a deconvolution problem,
\begin{equation*}
	\min_{H} ||H-X||_2^2+||F-\nabla H||^2_2.
\end{equation*}
It can be efficiently solved by the fast Fourier transform (fft) and inverse fft (ifft) operators, and the solution is
\begin{equation}\label{eq:H}
	H = {\rm ifft} \left\lbrace \frac{{\rm fft}(X)+\overline{{\rm fft}(k_h)}\odot{\rm fft}(F)}{1+\overline{{\rm fft}(k_h)}\odot{\rm fft}(k_h)} \right\rbrace,
\end{equation}
where $\overline{x}$ denotes the complex conjugation.

In order to make model more flexible, the hyper-parameters $\lambda$ and $\tau$ are automatically updated. According to empirical Bayes, we have
\begin{equation}\label{eq:hp1}
	\lambda = \frac{1}{hw}\sum_{i,j} \mathbb{E}[a_{ij}] = \frac{1}{hw}\sum_{i,j} \left( \frac{1}{\alpha_{ij}}+\frac{1}{\tilde{\lambda}}\right)
\end{equation}
and
\begin{equation}\label{eq:hp2}
\tau = \frac{1}{hw}\sum_{i,j}\mathbb{E}[b_{ij}] = \frac{1}{hw}\sum_{i,j} \left( \frac{1}{\beta_{ij}}+\frac{1}{\tilde{\tau}}\right).
\end{equation}

\subsection{Algorithm and implement details}
Algorithm \ref{alg:moal} summarizes the workflow of our proposed model, where E-step and M-step alternate with each other until the maximum iteration number $T^{\rm out}$ is reached. Since there is no analytic solution in M-step, we maximize $Q$-function by updating $(X,H,F)$ $T^{\rm in}$ times. It is found that $T^{\rm in}$ does not affect performance very much. To reduce computation, we set $T^{\rm in}=2$. Furthermore, it is found that algorithm generates a satisfactory result if the outer loop iterations $T^{\rm out}$ is set to 15. Note that hyper-parameter $\lambda_g$ and $\rho$ denote the strength of gradient and $L_2$ norm penalties, respectively. Empirical studies suggest to set $\lambda_g=0.5$ and $\rho = 0.001$.

\begin{algorithm}[h]
	\caption{Bayesian Fusion}
	\label{alg:moal}
	\begin{algorithmic}[1]
		\REQUIRE ~~\\ 
		Infrared image $U$, Visible image $V$, Maximum iteration number of outer and inner loops $T^{\rm out}$ and $T^{\rm in}$.
		\ENSURE ~~\\ 
		Fused image $I$.
		\STATE $Y=U-V$; Initialize $W_1,W_2=1,H,F=0,\lambda,\tau=1$;
		\FOR {$t=1,\cdots,T^{\rm out}$}
		\STATE \% (M-step)
		\FOR {$j=1,\cdots,T^{\rm in}$}
		\STATE Update $X,F,H$ with Eqs. (\ref{eq:X}), (\ref{eq:F}) and (\ref{eq:H}), respectively.
		\ENDFOR
		\STATE \% (E-step)
		\STATE Evaluate expectations by Eqs. (\ref{eq:E-step1}) and (\ref{eq:E-step2}).
		\STATE Update hyper-parameters $\lambda,\tau$ by Eqs. (\ref{eq:hp1}) and (\ref{eq:hp2}).
		\ENDFOR
		\STATE $I=X+V$.
	\end{algorithmic}
\end{algorithm}

\section{Experiments}\label{sec:2}
This section aims to study the behaviors of our proposed model and other popular counterparts, including  CSR\cite{liu2016image}, ADF\cite{bavirisetti2015fusion}, FPDE\cite{bavirisetti2017multi}, TSIFVS\cite{bavirisetti2016two} and TVADMM\cite{guo2017infrared}.
All experiments are conducted with MATLAB on a computer with Intel Core i7-9750H CPU@2.60GHz.

\subsection{Experimental data}
In this experiment, we test algorithm on TNO image fusion dataset\cite{TNO}\footnote{\url{https://figshare.com/articles/TNO Image Fusion Dataset/1008029}} and RGB-NIR Scene dataset\cite{brown2011multi}\footnote{\url{https://ivrlwww.epfl.ch/supplementary_material/cvpr11/index.html}}. 20 pairs of infrared and visible images in TNO dataset
and 52 pairs in the “country” scene of NIR dataset are employed. In TNO dataset, the interesting objects cannot be observed in visible images, as it was shot in night. In contrast, they are salient in infrared images, but without textures. While the NIR image dataset was obtained in daylight, and we test whether the fused image can have more detailed information and highlight information.

\subsection{Subjective visual evaluation}
\begin{figure}[!]
	\centering
	\includegraphics[width=1\linewidth]{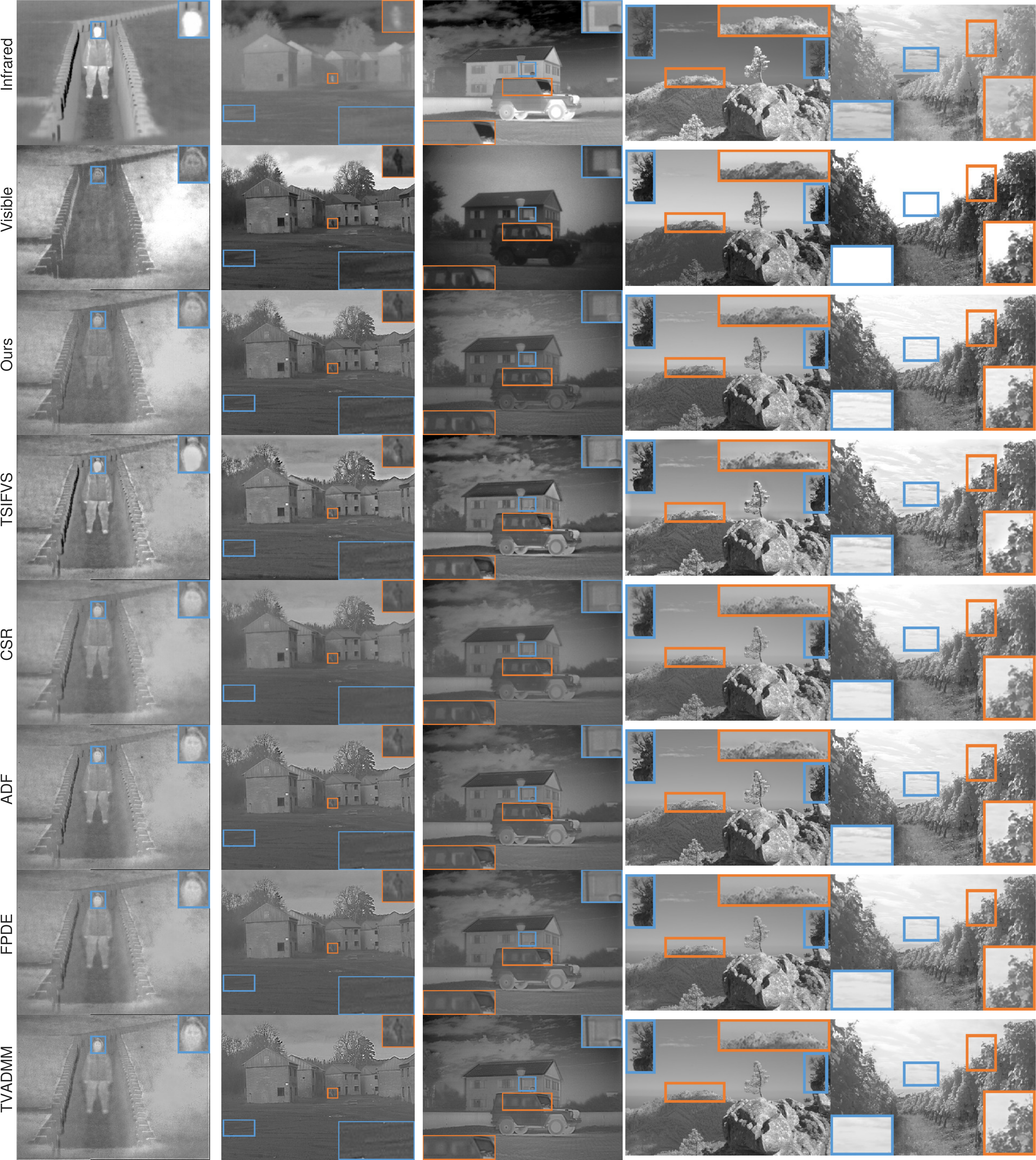}
	\caption{Qualitative fusion results.
From left to right:
``Soldier\_in\_trench 1'', ``Image\_04'' and ``Marne\_04'' in TNO dataset,
``Image\_13'' and ``Image\_35'' in NIR dataset.
From top to bottom: infrared images, visible images, results of our method, TSIFVS, CSR, ADF, FPDE and TVADMM methods.}
	\label{montage2035}
\end{figure}

In Figure \ref{montage2035}, the qualitative fusion results are exhibited, respectively.
From left to right:
``Soldier\_in\_trench\_1'', ``Image\_04'' and ``Marne\_04'' in TNO dataset,
``Image\_13'' and ``Image\_35'' in NIR dataset.
In the first column images, the TSIFVS and ADF methods have almost no face details. The TVADMM method has low target brightness, and the background of the CSR and FPDE methods (such as the trenches) is not clear enough.
Analysis of the fusion results for the second column images, apparently, the house details of the CSR method is poor and the ground detail of the ADF method is not obvious enough. Meanwhile, the target objective of the TVADMM and TSIFVS methods have low brightness, and the background details(e.g. the trees) of the FPDE method are not clear enough.
In the results of the third column images, the FPDE and ADF methods have lower brightness and fewer details, while the TVADMM and CSR methods have poorer window details, and the TSIFVS method has less obvious edge contours.
In the results of the fourth and fifth column images, the edge contour of the TSIVIF method does not fit the human visual system because of the clear boundary. The CSR and TVADMM methods are not salient enough in trees/clouds details and edges. Objects (trees and mountains) of the ADF method have poor highlighting effects and the FPDE method has visual blur with fewer details. 

In short, compared with the previous methods, our proposed Bayesian fusion model can generate better-fused images with high-light targets and rich texture details.

\subsection{Objective quantitative evaluation}
We calculate the average of the selected image pairs in Entropy (EN)\cite{roberts2008assessment}, Mutual information (MI)\cite{qu2002information}, $Q^{AB/F}$\cite{qu2002information}, Standard deviation (SD)\cite{rao1997fibre} and Structure similarity index measure (SSIM)\cite{wang2002universal} metrics for our proposed model and other popular counterparts. EN and SD measure how much information is contained in an image. $Q^{AB/F}$ reflects the edge information preserved in the fusion image. MI measures the agreement between source images and the fusion image, and SSIM reports the consistency in the light of structural similarities between fusion and source images. The larger metric values are, the better a fused image is. Please refer to \cite{ma2019infrared} to see more details on these metrics.

We show a quantitative comparison of these fusion methods in Table \ref{table}. In TNO dataset, our method performs best in terms of the MI, $Q^{AB/F}$, SD metrics, and is ranked second in the EN and SSIM indicators, in which the first are the TSIFVS and ADF methods. Meanwhile, in RGB-NIR Scene dataset, we get two first places in MI, SD and three second places in EN, $Q^{AB/F}$ and SSIM. This exhibition demonstrates the excellent performance of our method on infrared and visible image fusion compared with other image fusion methods.

\begin{table}[h]
	\centering
	\caption{Quantitative results of different methods. The largest value is shown in bold, and the second largest value is shown in underlined.}
\begin{tabular}{lcccccc}
\toprule
\multicolumn{7}{c}{\textbf{Dataset: TNO image fusion dataset}}\\
Metrics& TSIFVS&	TVADMM	&CSR&	ADF&	FPDE&	BayesFusion
 \\
 EN	&	\textbf{6.500} 	&	6.206 	&	6.225 	&	6.180 	&	6.255 	&	\underline{6.432}	\\
MI	&	1.649 	&	1.919 	&	1.900 	&	\underline{1.942 }	&	1.730 	&	\textbf{2.448}	\\
$Q^{AB/F}$	&	0.510 	&	0.340 	&	\underline{0.534 }	&	0.436 	&	0.508 	&	\textbf{0.549}	\\
SD	&	\underline{25.910 }	&	21.078 	&	21.459 	&	20.578 	&	21.327 	&	\textbf{26.285}	\\
SSIM	&	0.906 	&	0.905 	&	0.864 	&	\textbf{0.949 }	&	0.863 	&	\underline{0.937}	\\
 \midrule
 \multicolumn{7}{c}{\textbf{Dataset: RGB-NIR Scene Dataset}}\\
Metrics& TSIFVS&	TVADMM	&CSR&	ADF&	FPDE&	BayesFusion
 \\
EN	&	\textbf{7.300} 	&	7.129 	&	7.170 	&	7.105 	&	7.115 	&	\underline{7.201 }	\\
MI	&	3.285 	&	3.673 	&	3.699 	&	\underline{3.944} 	&	3.877 	&	\textbf{4.078} 	\\
$Q^{AB/F}$	&	0.571 	&	0.530 	&	\textbf{0.626 }	&	0.553 	&	0.580 	&	\underline{0.587 }	\\
SD	&	\underline{43.743} 	&	40.469 	&	40.383 	&	38.978 	&	39.192 	&	\textbf{46.105} 	\\
SSIM	&	1.157 	&	1.241 	&	1.130 	&	\textbf{1.274} 	&	1.249 	&	\underline{1.251} 	\\
\bottomrule
\label{table}
\end{tabular}%
\end{table}

\section{Conclusion}\label{sec:3}
In our paper, we present a novel Bayesian fusion model for infrared and visible images. In our model, the image fusion task is transformed into a regression problem, and a hierarchical Bayesian fashion is established to solve the problem. Additionally, the TV penalty is used to make the fused image similar to human visual system.
Then, the model is efficiently inferred by the EM algorithm with the half-quadratic splitting algorithm.
Compared with the previous methods in TNO and NIR datasets, our method can generate better fused images with highlighting thermal radiation target areas and abundant texture details, which can facilitate automatic detection and accurate positioning of targets.
\section*{Acknowledgements}
The research of S. Xu is supported by the Fundamental Research Funds for the Central Universities under grant number xzy022019059. The research of C.X. Zhang is supported by the National Natural Science Foundation of China under grant 11671317 and the National Key Research and Development Program of China under grant 2018AAA0102201. The research of J.M. Liu is supported by the National Natural Science Foundation of China under grant 61877049 and the research of J.S. Zhang is supported by the National Key Research and Development Program of China under grant 2018YFC0809001, and the National Natural Science Foundation of China under grant 61976174.
\bibliography{mybibfile}

\end{document}